\documentclass[10pt, a4paper]{article}

\usepackage{lrec2026}
\usepackage{booktabs}
\usepackage{multirow}

\title{From 124 Million Tokens to 1,021 Neologisms:\\ A Large-Scale Pipeline for Automatic Neologism Detection}

\name{Diego Rossini, Lonneke van der Plas}

\address{Università della Svizzera italiana (USI) \\
         Lugano, Switzerland \\
         \{diego.rossini, lonneke.vanderplas\}@usi.ch}

\abstract{
We present a scalable, modular pipeline for automatic neologism
detection that combines rule-based filtering with LLM classification.
The pipeline is grounded in two complementary word-formation
frameworks, grammatical and extra-grammatical morphology, which
jointly define the scope of what counts as a neologism and inform a
four-class classification scheme (\textsc{neologism}, \textsc{entity},
\textsc{foreign}, \textsc{none}). While designed to be modular and
transferable at the architectural level, the pipeline is instantiated
on 527 million English-language Reddit posts spanning 2005--2024.
From this corpus, we extract 124.6 million unique tokens and reduce
them by over 99.99\% to yield 1,021 neologism candidates, a set
small enough for manual expert verification. Multiple LLMs
independently classify each candidate via majority vote, with a
final verification step, revealing substantial cross-model
disagreement and highlighting the challenge of operationalizing
neologism detection at scale. Manual annotation of all 1,021
candidates confirms that 599 (58.7\%) are genuine lexical
innovations. The pipeline code, vocabulary compilation scripts, and
the annotated candidate list are available at
\url{https://github.com/DiegoRossini/neologism-pipeline}.
\\ \newline \Keywords{neologism detection, lexical innovation, Reddit,
large language models, rule-based filtering, computational neology}
}

\begin{document}

\maketitleabstract

\section{Introduction}
\label{sec:intro}

Although the study of neologisms has deep roots in linguistics
\citep{guilbert_creativite_1975, rey_neologisme_1976}, their automatic
detection is a comparatively recent task. Computational approaches only
became feasible once large machine-readable corpora were available in
the 1990s \citep{renouf_word_1993, cabre_strategie_1995}. Since then,
a number of web-based platforms have been developed for neologism
identification
\citep{kerremans_neocrawler_2012, cartier_neoveille_2017,
klosa_new_2018}. These systems typically depend on static
exclusion dictionaries and language-specific resources, and require
manual expert verification of their output
\citep{brasolin_ti_2023, cartier_neoveille_2017,
tomaszewska_neon_2025}. The key challenge for any detection pipeline is
therefore to reduce the candidate set to a size where such verification
is feasible.

More recently, social media data has attracted increasing attention as a
source for studying lexical innovation, given the volume, diversity, and
informality of user-generated content
\citep{grieve_mapping_2018, wurschinger_social_2021}. However, the same
characteristics that make these platforms attractive also pose a challenge
for neologism detection: in a large corpus, the vast majority of tokens
absent from standard dictionaries are not neologisms but typos,
misspellings, concatenated strings, code fragments, or foreign-language
material. In the dataset used in this study, 527 million
English-language Reddit posts spanning 2005--2024
\citeplanguageresource{baumgartner_pushshift_2020, watchful1_subreddit_2025}, we
extract 124.6 million unique tokens, which the pipeline reduces by over
99.99\% to yield 1,021 neologism candidates. Classifying each of the 124.6 million unique tokens individually with an LLM would be computationally infeasible,
which motivates a multi-stage filtering approach that progressively
narrows the set before classification.

In this paper, we present a pipeline for large-scale neologism detection
that combines deterministic rule-based filtering with LLM-based
classification. The rule-based stages progressively reduce the candidate
set; the LLM stage then classifies surviving candidates into four
categories: \textsc{entity}, \textsc{neologism}, \textsc{foreign}, or
\textsc{none}. Multiple LLMs independently classify each candidate, and
only those receiving a majority vote are retained. A final
verification step can then confirm or discard the output of the
preceding models. Our contributions are: (1)~a scalable, modular
pipeline for neologism detection from social media corpora, grounded in
word-formation theory (\S\ref{sec:theory}), whose output illustrates a range of grammatical and extra-grammatical word-formation processes (\S\ref{sec:discussion:wf}); (2)~a comparative
evaluation of multiple LLMs on a four-class neologism classification
task; and (3)~a detailed manual analysis of all 1,021 pipeline output candidates, including gold annotation, error analysis by category (\S\ref{sec:discussion:errors}), and classification of detected neologisms along word-formation processes (\S\ref{sec:discussion:wf}).\footnote{The pipeline code, vocabulary compilation scripts, and the annotated candidate list are available at \url{https://github.com/DiegoRossini/neologism-pipeline}.}

\section{Related Work}
\label{sec:related}

The dominant paradigm for automatic neologism detection remains the
\emph{exclusion dictionary method}: a token is flagged as a candidate
neologism if it does not appear in one or more reference lexicons
\citep{renouf_word_1993, cabre_strategie_1995}. This principle
underpins the major detection platforms developed over the past two
decades, including the NeoCrawler
\citep{kerremans_neocrawler_2012, kerremans_using_2018}, which
monitored English-language websites for previously unattested forms,
N\'{e}oveille \citep{cartier_neoveille_2017}, which adopted a similar
architecture for multiple languages, and the IDS
Neologismenw\"{o}rterbuch \citep{klosa_new_2018}, a continuously
updated German neologism dictionary backed by corpus monitoring. The
NeoCrawler was formally decommissioned in 2020
(Q.~W\"{u}rschinger, personal communication, 2025), illustrating the
fragility of long-term tool availability. All three systems depend on
language-specific resources that limit portability across languages
and corpora. Our pipeline adopts the same foundational exclusion principle but separates language-specific resources from the architectural design: the sequence of filtering stages is pre-determined, while the resources they operate on (reference vocabularies, phonotactic rules, frequency dictionaries) should be substituted or adapted for each target language.

Beyond pure exclusion lookup, \citet{falk_non_2014} trained an SVM on
French newspaper candidates using form-related, morpho-lexical, and
thematic features, demonstrating the value of semantic context for
neologism classification. Our pipeline follows a similar two-stage
logic, but delegates the classification step to prompted LLMs rather
than to feature-based classifiers.

As corpora drawn from social media have grown in size, so has the
difficulty of the candidate extraction step itself.
\citet{grieve_mapping_2018} identified 54 emerging words from 8.9
billion tokens of geolocated American Twitter data by tracking
frequency increases over time and filtering manually.
\citet{mahler_lexical_2020} applied a comparable frequency-based
methodology to Reddit, and \citet{wurschinger_social_2021}
demonstrated how network metrics capture properties of lexical
innovation that frequency measures alone cannot reveal. \citet{brasolin_ti_2023} and
\citet{spina_detecting_2024} extracted candidates from millions of
geolocated Italian tweets through exclusion filtering and manual
distillation, identifying hundreds of unattested word forms. A
recurring challenge across these studies is that the vast majority of
tokens absent from standard dictionaries are not neologisms but typos,
misspellings, code fragments, or foreign-language material.

The most directly comparable recent system is NeoN
\citep{tomaszewska_neon_2025}, a multi-layered pipeline for Polish
that combines frequency analysis, structural constraints, reference
corpus checking, and spelling error detection with an LLM-based final
filter, demonstrating that LLMs can serve as effective precision
boosters after rule-based pre-filtering. Our work differs from NeoN in several respects: we ground the pipeline in two complementary word-formation frameworks (\S\ref{sec:theory}) that define the scope of each category; we adopt a four-class taxonomy rather than a binary filter; we use a multi-model majority-vote scheme with independent verification rather than a single LLM; and we provide a detailed manual analysis of all pipeline output, including error categorisation and classification by word-formation process.

\section{Theoretical Foundations}
\label{sec:theory}

Any neologism detection pipeline presupposes an operational definition
of what counts as a neologism. This section presents the two
word-formation frameworks that jointly inform the design of our
classification scheme and, in particular, determine the scope of the
\textsc{neologism} label assigned by the LLM stage
(\S\ref{sec:method}).

\subsection{Grammatical Word Formation}
\label{sec:theory:stekauer}

\citeauthor{stekauer_onomasiological_1998}'s onomasiological theory
\citep{stekauer_onomasiological_1998,
stekauer_fundamental_2001, den_dikken_onomasiological_2005} models
word formation as a top-down, need-driven process: a speaker identifies
a concept lacking a conventional expression and coins a new naming unit
by selecting an onomasiological type --- a structural pattern that maps
conceptual content onto morphological form. The process is
\textit{grammatical} in the sense that, given a naming need, the
resulting formation is constrained by the productive onomasiological
types available in the language.

A central consequence of this framework concerns nonce-formations.
Against the view that these are deviant, context-dependent, and
inherently non-lexicalisable coinages
\citep[cf.][]{hohenhaus_non-lexicalizability_1998},
\citet{stekauer_theory_2002} argues that nonce-formations are regular
products of the Word-Formation Component, generated by the same
productive rules as any other naming unit. What distinguishes them is
not structural deviance but \emph{lifecycle stage}: a nonce-formation
is a neologism at the earliest point between coinage and
dissemination, and whether it subsequently becomes institutionalised or
falls out of use is an empirical matter that cannot be predicted at
the time of coining. The notion of ``nonce-formation'' as a
structurally distinct category thus collapses into a temporal label.

For the pipeline, this entails that tokens attested infrequently in
the corpus cannot be excluded as non-neologisms on formal grounds
alone, since nonce-formations are structurally indistinguishable from
formations that will eventually become established. The frequency
threshold (\S\ref{sec:method:freq}) is accordingly designed to filter
noise rather than to impose a lexicalisation requirement. However,
\citeauthor{stekauer_theory_2002}'s model is explicitly limited to
rule-governed formation: processes such as blending, clipping, and
acronymy, whose output cannot be derived from productive
onomasiological types, fall outside the Word-Formation Component and
are relegated to the Lexicon
\citep{stekauer_fundamental_2001}. The following framework addresses
precisely this gap.

\subsection{Extra-Grammatical Word Formation}
\label{sec:theory:mattiello}

\citet{mattiello_extra-grammatical_2013} addresses those processes
that fall outside the scope of grammatical morphology: clippings,
blends, acronyms, abbreviations, and other formations whose input
does not allow prediction of a regular output through any rule-based
model of word formation. Within the framework of Natural Morphology
\citep{dressler_extragrammatical_2000}, these are classified as
\textit{extra-grammatical}, distinct from both core grammatical
morphology (rule-governed, productive) and marginal morphology
(partially regular). Although traditionally marginalised for their
irregularity and unpredictability,
\citet{mattiello_extra-grammatical_2013} demonstrates that
extra-grammatical processes are productive in their own right,
particularly in informal registers, and that they comply with criteria
of well-formedness and contextual suitability. The driving mechanism is \textit{analogy} rather than rule application
\citep{mattiello_extra-grammatical_2013, mattiello_analogical_2017,
arndt-lappe_word-formation_2015}: new formations arise by modelling
on existing words, either individually (surface analogy) or through
recurrent patterns (analogy via schema).
\citet{mattiello_analogical_2017}, following
\citet{booij_construction_2010} and
\citet{plag_morphological_1999}, extends this mechanism to playful
coinages such as \textit{doggo} (← \textit{dog}), previously excluded as \textit{expressive morphology}, i.e.\ playful or affective modifications of existing words \citep{zwicky_plain_1987}. Moreover, the boundary
between extra-grammatical and grammatical morphology is not fixed, as
formations that originate as creative coinages can over time become
regular and productive \citep{kortvelyessy_creativity_2022,
kortvelyessy_role_2021}.

For the pipeline, this entails that the \textsc{neologism} class must
be broad enough to encompass both grammatical and extra-grammatical
formations. Taken together, the two frameworks define the theoretical
scope of the positive labels used in this study: the four-class
classification scheme presented in \S\ref{sec:method}
operationalises this joint definition, with the \textsc{neologism} and
\textsc{entity} classes capturing genuine lexical innovations and
\textsc{foreign} and \textsc{none} isolating non-neologistic material
that rule-based filtering alone cannot eliminate.

\section{Methodology}
\label{sec:method}

The pipeline is designed to be modular: the sequence of filtering stages is pre-determined, but the resources each stage operates on (reference vocabularies, phonotactic rules, frequency dictionaries) are language-specific and must be substituted for each target language. The instantiation described below targets English.

\subsection{Tokenization}
\label{sec:method:tokenization}

Raw texts are tokenized using a spaCy language model, with named
entity recognition, dependency parsing, and lemmatization disabled
for efficiency. The choice of model depends on the target language.
A corpus-specific preprocessing step removes or replaces non-lexical
content before tokenization: for social media corpora, this may
include URLs, platform-specific references, user mentions, hashtags,
emojis, and non-ASCII characters; for other corpus types, different
noise patterns (e.g., markup tags, metadata fields) may require
analogous treatment. Punctuation, stopwords, and whitespace tokens
are discarded, and all remaining tokens are lowercased.

\subsection{Vocabulary Filtering}
\label{sec:method:vocabfilter}

Tokens present in a reference vocabulary compiled exclusively from
sources predating a chosen cutoff date are filtered out on the
assumption that they represent established lexical items rather than
neologisms. The cutoff defines an observation window over which the
lifecycle of detected neologisms can be tracked. The pipeline accepts
one or more reference vocabularies; when multiple sources are
available, combining them reduces the risk of false positives caused
by gaps in any individual lexicon. The composition can be tailored to
the target language and corpus: for social media data, it may include
platform-specific vocabulary, crowdsourced slang dictionaries, and
encyclopaedic entries alongside standard lexicons, while for more
formal corpora, curated dictionaries and domain-specific
terminologies may suffice. Any token found in the reference
vocabulary is excluded from further processing.

\subsection{Pattern-based Cleaning}
\label{sec:method:pattern}

Tokens surviving the vocabulary filter are subjected to pattern-based
rules designed to remove noise that no dictionary would capture. These
rules fall into two categories. The first is language-independent:
tokens must be purely alphabetic and fall within a configurable length
range, while those exhibiting excessive character repetition, low
character entropy, or repeated character sequences are discarded as
likely keyboard spam or encoding artefacts. The second category is
language-specific and must be adapted to the target language: this
includes phonotactic constraints (e.g., implausible consonant or vowel
clusters), expressive spelling variants (e.g., elongated interjections,
laughter patterns), and corpus-specific noise patterns (e.g.,
placeholder or template text).

\subsection{Typo and Concatenation Detection}
\label{sec:method:typo_concat}

Corpora, particularly those drawn from social media, frequently
contain misspellings and tokens formed by words concatenated without
spaces. Neither constitute lexical innovations, yet both survive
vocabulary filtering because they do not match any reference entry.
The pipeline applies SymSpell \citeplanguageresource{symspell}, a symmetric delete
spelling correction algorithm with support for multiple languages,
to detect both cases against a reference frequency dictionary. A
token is flagged as a typo if it falls within a configurable edit
distance of a high-frequency entry in the dictionary, and as a
concatenation if it can be segmented into two or more parts each
appearing in the same dictionary. Minimum character length thresholds prevent spurious
matches on short tokens, and a conservative
maximum edit distance ensures that only tokens
closely resembling high-frequency dictionary entries
are flagged, so that morphologically complex forms such as compounds or blends are unlikely to be flagged; those that are can be recovered by the frequency-based reintegration mechanism (\S\ref{sec:method:freq}).

\subsection{Frequency Threshold and Reintegration}
\label{sec:method:freq}

Tokens previously excluded as typos or concatenations
(\S\ref{sec:method:typo_concat}) are reconsidered if they meet
a configurable frequency threshold. If a token resembles a misspelling or a segmentable string yet recurs frequently in the corpus, it is unlikely to be an error, and its reintegration prevents genuine coinages from being prematurely discarded.

Candidates occurring fewer than the frequency threshold are
excluded. While this introduces a tension with the theoretical
position outlined in \S\ref{sec:theory:stekauer}, where
nonce-formations are treated as legitimate neologisms regardless of
their frequency, the constraint is computational rather than
theoretical: when too many candidates survive the rule-based filters,
manual or LLM-based verification becomes infeasible. The threshold
can be adjusted or omitted entirely depending on corpus size and
available resources. In practice, nonce-formations attested below
the threshold are lost; however, the threshold is not designed to
impose a lexicalisation requirement but to separate deliberate,
repeated use from accidental variation, since typos and random
strings rarely recur at scale.

\subsection{Foreign Language Detection}
\label{sec:method:foreign}
Depending on the target language, the corpus may contain substantial
material from other languages that survives vocabulary filtering.
The pipeline applies the Lingua language detector
\citeplanguageresource{stahl_lingua} to flag and filter tokens
identified as belonging to a language other than the target. A
configurable confidence threshold controls how aggressively tokens
are filtered. Tokens whose confidence score falls below the
threshold, for instance due to mixed Tagalog--English morphology or
orthographic overlap with the target language, are retained and
delegated to the LLM classification stage, which includes a
dedicated \textsc{foreign} category. The stage does not distinguish foreign-language noise from loanwords entering the target language; this limitation is discussed in the Limitations section.

\subsection{LLM Classification}
\label{sec:method:llm}
Candidates surviving the filtering stages are classified into four
categories using large language models. The taxonomy reflects the
theoretical scope established in \S\ref{sec:theory}:
\textsc{neologism} (new words, slang, or words derived from proper
nouns, encompassing both grammatical and extra-grammatical
formations); \textsc{entity} (proper nouns such as people, companies,
brands, products, or places); \textsc{foreign} (words from other
languages that escaped the language detection stage,
\S\ref{sec:method:foreign}); and \textsc{none} (residual noise
including usernames, typos, programming terms, and unclear cases).

Both \textsc{neologism} and \textsc{entity} constitute lexical
innovations: in \citeauthor{stekauer_onomasiological_1998}'s
onomasiological framework, word-formation is a naming response to
newly salient extra-linguistic referents, and proper nouns denoting
emerging social or cultural entities qualify as newly established
naming units. The two classes are kept separate for analytical
purposes, facilitating comparison with standard NER categories in
downstream applications.
Classification is performed in two stages. First, multiple LLMs
independently classify each candidate token. Labels are aggregated
via majority vote: a token receives a label only if the majority of
models agree; otherwise it is marked \textsc{unknown}. Second, an
additional model verifies each label and produces the final output.
The choice and number of models is configurable; using multiple architectures trained on different data reduces idiosyncratic misclassifications, while the independent verification step provides an additional quality control layer at minimal additional cost.

\section{Experimental Setup}
\label{sec:setup}

This section describes the instantiation of the pipeline for
English-language neologism detection on Reddit data. All
language-specific resources, parameters, and model choices reported
below can be substituted for other languages or corpora.

\subsection{Corpus}
\label{sec:setup:corpus}

The corpus consists of Reddit submissions and comments spanning
January 2005 to December 2024, drawn from the Pushshift archive
\citeplanguageresource{baumgartner_pushshift_2020, watchful1_subreddit_2025}. After
excluding deleted posts, removed content, and non-textual submissions,
the dataset comprises approximately 527 million posts. Although the
corpus is predominantly English, it contains multilingual content,
most notably Taglish (Tagalog--English code-switching) in
Filipino-oriented subreddits, as well as posts in Portuguese, Spanish,
French, German, and other languages.

\subsection{Tokenization}
\label{sec:setup:tokenization}

We use spaCy's \texttt{en\_core\_web\_lg} model with named entity
recognition, dependency parsing, and lemmatization disabled for
efficiency. URLs, subreddit references (\texttt{r/\textbackslash w+}),
user mentions (\texttt{u/\textbackslash w+}), and hashtags are replaced
with placeholder tokens; emojis and non-ASCII characters are removed.
Punctuation, stopwords, and whitespace tokens are discarded, and all
remaining tokens are lowercased. The tokenization stage yields 124.6
million unique token types.

\subsection{Reference Vocabularies}
\label{sec:setup:vocab}

The reference vocabulary is compiled exclusively from pre-2015
sources, establishing a ten-year observation window (2015--2024) over
which newly emerged tokens can be identified. The combined vocabulary
comprises 16.3 million unique surface forms drawn from six sources
(Table~\ref{tab:setup}). Using multiple independently compiled
resources reduces the risk of false positives caused by gaps in any
individual lexicon: Reddit and Urban Dictionary \citeplanguageresource{urban_dictionary} cover informal
register, Wikipedia titles \citeplanguageresource{wikimedia_wikipedia_2015} capture named entities and technical
terminology, while WordNet \citeplanguageresource{princeton_wordnet}, Wiktionary \citeplanguageresource{wikimedia_wiktionary_2015}, and NoSlang (5.5K tokens, obtained with permission from the site owner) provide baseline
lexical coverage.

\subsection{Filtering Parameters}
\label{sec:setup:filtering}

\paragraph{Pattern cleaning.}
Tokens must be purely alphabetic and between 3 and 20 characters in
length. The English-specific rules filter tokens starting with double
vowels (\textit{aa}, \textit{ee}, \textit{ii}, \textit{oo},
\textit{uu}), implausible consonant clusters, expressive variants
(\textit{hahaha}, \textit{yeaah}, \textit{ughh}), repeated character
sequences, and Lorem Ipsum placeholder words. Tokens exceeding six
characters with two or fewer unique characters are discarded as
keyboard spam. The full rule set is available in the project repository.

\paragraph{Typo and concatenation detection.}
SymSpell \citeplanguageresource{symspell} is configured with a maximum edit distance
of 2 and a frequency dictionary compiled from Reddit pre-2015 token
frequencies and WordNet. A token is flagged as a typo if its closest
match in the dictionary has edit distance 1--2 and frequency above
100; minimum token length for typo checking is 5 characters.
Concatenation detection applies word segmentation on tokens of at
least 6 characters, flagging those that segment into two or more parts all present in the frequency dictionary. Genuine compounds flagged at this stage can be recovered by the reintegration mechanism described below. 

\paragraph{Frequency threshold.}
The minimum occurrence threshold is set to 100
(\S\ref{sec:method:freq}). Tokens previously flagged as typos or
concatenations are reintegrated if they meet this threshold.

\paragraph{Foreign language detection.}
The Lingua language detector (Stahl, 2022) is applied with a confidence threshold of 0.75 across 47 languages, removing 33,959 tokens (16.3\% of the 208,932 candidates at that stage).

\medskip
\noindent All thresholds reported above were set based on preliminary
experimentation; a discussion of their limitations is provided
in the Limitations section.

\subsection{LLM Classification}
\label{sec:setup:llm}

The three open-source models---Qwen 2.5 72B, LLaMA 3.3 70B, and Mistral Large 2 123B---independently classify each candidate; labels
are aggregated via majority vote (\S\ref{sec:method:llm}). Claude 4.5
Haiku serves as an independent verification source and does not
participate in the vote.

\begin{table}[t]
\centering
\small
\begin{tabular*}{\columnwidth}{@{\extracolsep{\fill}}lrl@{}}
\toprule
\multicolumn{3}{@{}l}{\textbf{Reference Vocabularies (all pre-2015)}} \\
\midrule
\textbf{Source} & \textbf{Tokens} & \textbf{Coverage} \\
\midrule
Reddit pre-2015 & 10.5M & Informal, platform jargon \\
Wikipedia titles & 4.4M & Entities, technical terms \\
Urban Dictionary & 1.5M & Slang \\
Wiktionary & 554K & Morphological variants \\
WordNet 3.1 & 147K & Core vocabulary \\
NoSlang & 5.5K & Chat abbreviations \\
\midrule
\textbf{Total} & \textbf{16.3M} & \\
\bottomrule
\end{tabular*}
\caption{Reference vocabularies and primary coverage.}
\label{tab:setup}
\end{table}

All open-source models use few-shot prompting with eight labelled
examples spanning the four classes and up to three contextual
sentences per candidate, drawn from diverse subreddits. Claude 4.5
Haiku classifies tokens with the same contextual examples as the
other models. The full prompt templates are provided in Appendix~\ref{sec:appendix:prompts}.

\subsection{Computational Setup}
\label{sec:setup:infra}

All experiments were run on a single multi-GPU server with
500~GB RAM and 4 GPUs (120~GB each). The open-source models were sharded across all four GPUs in bfloat16 precision. Claude 4.5
Haiku was accessed via the Anthropic Batch API.

On the described hardware, the ideal critical path is
approximately 50--65 hours ($\sim$2--3 days): tokenization of
527 million posts accounts for 18--24 hours, vocabulary filtering
and context retrieval for $\sim$9 hours, and sequential LLM
inference over three models for 22--30 hours (40--50\% of total
compute). Running the three models in parallel on separate nodes
would reduce the total to $\sim$38--49 hours.

\section{Results}
\label{sec:results}

\subsection{Filtering Cascade}
\label{sec:results:cascade}

Table~\ref{tab:cascade} reports the number of candidate tokens
surviving each pipeline stage. The rule-based stages reduce the
initial 124.6 million unique tokens by 99.86\%, yielding 174,973
candidates for LLM classification. The most aggressive single stage
is pattern cleaning, which removes 90 million tokens (72.2\% of the
input at that point), followed by concatenation detection (13.2 million
concatenated tokens) and vocabulary lookup (10.7 million known
words). The frequency threshold eliminates a further 6.9 million
low-frequency tokens. Of the tokens previously excluded as typos or
concatenations, 118,544 meet the frequency threshold and are
reintegrated into the candidate pool (\S\ref{sec:method:freq}).

\begin{table}[t]
\centering
\small
\begin{tabular*}{\columnwidth}{@{\extracolsep{\fill}}lr@{}}
\toprule
\textbf{Stage} & \textbf{Remaining} \\
\midrule
Tokenization & 124,593,754 \\
Vocabulary lookup & 113,909,871 \\
Pattern cleaning & 23,955,763 \\
Concatenation detection & 10,793,055 \\
Typo detection & 7,065,796 \\
Freq.\ threshold + reintegration & 208,932 \\
Foreign language detection & 174,973 \\
\midrule
Majority vote (\textsc{neologism}) & 10,499 \\
Haiku verification & 1,021 \\
\bottomrule
\end{tabular*}
\caption{Filtering cascade: candidates remaining after each stage.}
\label{tab:cascade}
\end{table}

\subsection{LLM Classification and Inter-Model Agreement}
\label{sec:results:llm}

The three open-source models independently classified all 174,973
tokens. Table~\ref{tab:model_dist} reports their label distributions,
revealing substantial cross-model disagreement. LLaMA is the most
aggressive \textsc{neologism} predictor (12.2\%, nearly double the
other two models), while Mistral is the most conservative overall,
assigning \textsc{none} to 59.9\% of tokens. Qwen detects the most
foreign-language material (22.2\%).

Unanimous agreement across all three models is reached for only
45.8\% of tokens (80,220); 48.4\% are decided by a 2-out-of-3
majority, and 5.8\% (10,134) result in three-way ties,
conservatively resolved to \textsc{none}. These complementary biases
validate the ensemble design: no single model would achieve the same
coverage.

The majority vote produces 10,499 \textsc{neologism} candidates
(6.0\%), 47,276 \textsc{entity} (27.0\%), 33,159 \textsc{foreign}
(19.0\%), and 84,039 \textsc{none} (48.0\%).

\subsection{Haiku Verification}
\label{sec:results:haiku}
Claude 4.5 Haiku independently classified the same tokens with the
same contextual examples. Applied as a verification filter to the
10,499 majority-vote \textsc{neologism} candidates, Haiku confirmed
897 as \textsc{neologism} (8.5\%), relabeled 124 as \textsc{entity}
(1.2\%), and rejected 9,478 to \textsc{none} (90.3\%). This high
rejection rate is driven primarily by model conservatism rather than
prompt design or category confusion. Haiku receives the same
multi-context prompts as the open-source models, yet assigns
\textsc{entity} to only 124 of 174,973 tokens (0.07\%), compared to
47,276 from the majority vote, effectively defaulting all uncertain
cases to \textsc{none} as instructed by the prompt. Across all
tokens, it assigns \textsc{none} to 89.4\%. This pattern places
Haiku at the extreme end of a conservatism spectrum already visible
among the open-source models, where Mistral (59.9\% \textsc{none}) is
markedly more conservative than Qwen (37.9\%) and LLaMA (37.7\%).
Table~\ref{tab:model_dist} reports the full label distribution
across all four models and the majority vote. The most striking
pattern is Haiku's near-total rejection of the \textsc{entity}
class: of 47,276 majority-vote entities, none are confirmed and
97.3\% are relabeled \textsc{none}.
The verification stage thus acts as a strict precision filter,
reducing the candidate set from 10,499 to 1,021.

\begin{table*}[t]
\centering
\small
\begin{tabular*}{\textwidth}{@{\extracolsep{\fill}}lrrrrr@{}}
\toprule
\textbf{Label} & \textbf{Qwen 72B} & \textbf{Mistral 123B} & \textbf{LLaMA 70B} & \textbf{Maj.\ vote} & \textbf{Haiku} \\
\midrule
\textsc{neologism} & 13,661 (7.8\%) & 11,493 (6.6\%) & 21,353 (12.2\%) & 10,499 (6.0\%) & 897 (0.5\%) \\
\textsc{entity} & 56,144 (32.1\%) & 32,311 (18.5\%) & 55,088 (31.5\%) & 47,276 (27.0\%) & 124 (0.1\%) \\
\textsc{foreign} & 38,851 (22.2\%) & 26,441 (15.1\%) & 32,625 (18.6\%) & 33,159 (19.0\%) & 17,506 (10.0\%) \\
\textsc{none} & 66,317 (37.9\%) & 104,728 (59.9\%) & 65,907 (37.7\%) & 84,039 (48.0\%) & 156,446 (89.4\%) \\
\bottomrule
\end{tabular*}
\caption{Label distribution per model and majority vote across all 174,973 tokens (count and \% of total). LLaMA's \textsc{none} count includes 954 unparseable responses.}
\label{tab:model_dist}
\end{table*} 

\subsection{Gold Standard Evaluation}
\label{sec:results:gold}
The first author manually annotated all 1,021 pipeline output
candidates using the same four-class taxonomy, following the
annotation criteria derived from the theoretical framework in
\S\ref{sec:theory}: tokens were classified as \textsc{neologism} if
they resulted from a word-formation process (grammatical or
extra-grammatical in the sense of
\citealt{mattiello_extra-grammatical_2013}) and were first attested
after 2015, as \textsc{entity} if they denoted a proper noun first
attested after 2015, as \textsc{foreign} if they belonged to another
language, and as \textsc{none} otherwise.
Table~\ref{tab:gold} cross-tabulates pipeline output against gold
labels.

\begin{table}[t]
\centering
\small
\begin{tabular*}{\columnwidth}{@{\extracolsep{\fill}}lrr@{}}
\toprule
\textbf{Gold label} & \textbf{Count} & \textbf{\%} \\
\midrule
Lexical innovation & 599 & 58.7 \\
\quad \textit{of which} \textsc{neologism} & 465 & 45.5 \\
\quad \textit{of which} \textsc{entity} & 134 & 13.1 \\
Non-neologism & 422 & 41.3 \\
\quad \textit{of which} \textsc{foreign} & 61 & 6.0 \\
\quad \textit{of which} \textsc{none} & 361 & 35.4 \\
\midrule
\textbf{Total} & \textbf{1,021} & \textbf{100} \\
\bottomrule
\end{tabular*}
\caption{Gold annotation of the 1,021 pipeline output candidates.}
\label{tab:gold}
\end{table}

Of the 1,021 candidates, 599 (58.7\%) are genuine lexical
innovations: 465 neologisms and 134 named entities. The remaining
422 consist of 361 false positives (\textsc{none}) and 61
foreign-language tokens that escaped both rule-based and LLM-based
detection.

\section{Discussion}
\label{sec:discussion}

The pipeline is best understood as a high-recall candidate generator
rather than a precision classifier. Its primary contribution is the
122,031:1 compression ratio, which reduces a task that no human
annotator could feasibly undertake (reviewing 124.6 million tokens)
to one that a single annotator can complete (reviewing 1,021
candidates). We do not report corpus-level recall, as the gold standard
covers only the pipeline output; the number of neologisms in the
124.6 million tokens that the pipeline may have missed is unknown.
To give a rough idea, however, an estimate based on an external
reference list is provided in \S\ref{sec:discussion:fn}. This framing aligns with how
comparable systems are evaluated:
\citet{tomaszewska_neon_2025} report precision at each stage and
note that recall is not computable;
\citet{grieve_mapping_2018} and \citet{brasolin_ti_2023} similarly
report counts of emerging words found. As in those systems, a final
manual verification step is an integral part of the design, not a
limitation.

\subsection{Word-Formation Patterns}
\label{sec:discussion:wf}

The 599 gold lexical innovations exhibit a range of word-formation
processes that connect directly to the theoretical frameworks in
\S\ref{sec:theory}. Table~\ref{tab:wf} organises the most
productive patterns along two axes: whether the formation is
analogical or non-analogical, and whether the process is
grammatical, marginal, or extra-grammatical in the sense of
\citet{mattiello_extra-grammatical_2013}.

\begin{table}[t]
\centering
\small
\begin{tabular}{@{}p{0.48\columnwidth}p{0.44\columnwidth}@{}}
\toprule
\textbf{Process} & \textbf{Examples} \\
\midrule
\multicolumn{2}{@{}l}{\textit{Analogical formations}} \\[0.3em]
Extra-gramm.: surface analogy & \textit{updoot, pawrents} \\
Extra-gramm.: analogy via schema & \\
\quad Secreted c.f.: \textit{-fluencer} & \textit{finfluencer, fitfluencer} \\
\quad Abbreviated c.f.: \textit{trad-} & \textit{tradwife, tradfem} \\
\midrule
\multicolumn{2}{@{}l}{\textit{Non-analogical formations}} \\[0.3em]
Grammatical: prefixation & \textit{deplatform, exvegan} \\
Grammatical: suffixation & \textit{wokeism, trumpism} \\
Grammatical: compounding & \textit{deepfake, longcovid} \\
Marginal: neoclassical c.f. & \textit{abrosexual, acephobia} \\
Extra-gramm.: blending & \textit{barbenheimer, maskne} \\
Extra-gramm.: expressive morph. & \textit{thiccest, consoomer} \\
\bottomrule
\end{tabular}
\caption{Word-formation processes among gold neologisms.}
\label{tab:wf}
\end{table}

Among non-analogical formations, standard grammatical processes
account for a substantial share of the data. Prefixation with
productive English prefixes yields forms such as \textit{deplatform},
\textit{detrash}, \textit{exvangelical}, and \textit{exvegan}.
Suffixation with \textit{-ism} generates \textit{wokeism},
\textit{trumpism}, \textit{defaultism}, and \textit{longtermism}.
Compounding produces \textit{deepfake}, \textit{longcovid}, and
\textit{vibecheck}. At the margins of grammatical morphology,
neoclassical combining forms of Latin or Greek origin appear in
novel bases: \textit{-sexual} (\textit{abrosexual},
\textit{dreamsexual}) and \textit{-phobia/-phobic}
(\textit{acephobia}, \textit{enbyphobic}). Non-analogical
extra-grammatical processes include blending, where two source words
are fused without following a prior model (\textit{barbenheimer},
\textit{maskne}, \textit{trumpanzee}), and expressive morphology,
where deliberate phonological distortion of existing words produces
new forms (\textit{thiccest}, \textit{consoomer}, \textit{chonkster}).

The analogical formations are exclusively extra-grammatical. Surface
analogy, where a single word serves as model, accounts for cases
such as \textit{updoot} (after \textit{upvote}) and
\textit{pawrents} (after \textit{parents}). More productive are
formations arising through analogy via schema, where a recurrent
fragment extracted from an initial blend becomes a combining form
used across a series. Several such combining forms have undergone
semantic generalisation, or secretion in the terminology of
\citet{mattiello_extra-grammatical_2013}: \textit{-fluencer} (from
\textit{influencer}; \textit{finfluencer}, \textit{fitfluencer},
\textit{scamfluencers}), \textit{-cel} (from \textit{incel};
\textit{femcel}, \textit{mentalcels}), \textit{-core} (from
\textit{hardcore}; \textit{goblincore}, \textit{traumacore}),
\textit{-nomics} (from \textit{economics}; \textit{bidenomics},
\textit{tokenomics}), \textit{-pilled} (from \textit{redpilled};
\textit{blackpilled}, \textit{blackpillers}), and
\textit{-maxxing} (from \textit{maxing}; \textit{looksmaxxing},
\textit{gymmaxxing}). Others function as abbreviated combining
forms without semantic reinterpretation: \textit{trad-} (from
\textit{traditional}; \textit{tradwife}, \textit{tradfem}) and
\textit{-flation} (from \textit{inflation}; \textit{greedflation},
\textit{pissflation}). Whether the secreted forms have fully
detached from their source words or remain at an intermediate stage
between splinter and combining form is a diachronic question that
the present data cannot resolve.

\subsection{Error Analysis}
\label{sec:discussion:errors}

The 422 non-neologistic tokens in the output fall into distinct
categories, each pointing to a specific pipeline limitation.

\paragraph{False positives (361 tokens).}
The largest error category comprises concatenations
(two or more words typed without a space, a common Reddit
orthographic artifact) such as \textit{datingapp},
\textit{sidehustle}, and \textit{telegramchannel} (approximately
50 tokens). These were correctly identified and removed by the
word segmentation step, but subsequently reintegrated by the
frequency threshold mechanism (\S\ref{sec:method:freq}), which
restores all tokens with $\geq$100 occurrences regardless of the
reason for their exclusion. At that point, the LLMs should have
classified them as \textsc{none}, but failed to recognise them as
mere orthographic variants of existing word sequences.
A second cluster (approximately 50 tokens) comprises tokens from
gaming and technical domains marked \textsc{none} for heterogeneous
reasons. Some are names of programming functions or UI components
(\textit{floatlayout}, \textit{floatmenu}, \textit{floattensor}):
accepting these would entail treating entire programming language
vocabularies as natural language neologisms. Others are spaceless
concatenations of pre-existing proper names (\textit{bioniccommando},
a 1987 Capcom title; \textit{biorepeel}, a cosmetic brand). Still
others are fragments of game-internal proper names where the pipeline
captured only part of a multi-token entity, or tokens that predate
2015 but were too domain-specific for the reference vocabulary.
Approximately 30 tokens are misspellings of neologisms themselves
(\textit{neurodivegent}, \textit{dollfication},
\textit{nuerotypicals}): the typo filter checks edit distance
against \emph{dictionary} words only and cannot detect that
\textit{neurodivegent} is a misspelling of \textit{neurodivergent},
which is itself absent from the reference vocabulary.
Finally, approximately 25 tokens are pre-2015 words absent from
the 16.3 million-word reference vocabulary (\textit{latinx},
\textit{onfleek}, \textit{biliteracy}): the vocabulary is
comprehensive but not exhaustive for informal and slang registers.

\paragraph{Foreign language leakage (61 tokens).}
Two patterns dominate. Taglish (Tagalog--English) code-switching accounts for 29 of 61 tokens (47.5\%): Tagalog prefixes (\textit{na-}, \textit{naka-}, \textit{sina-}) affixed to English roots (\textit{naghost}, \textit{nakablock}, \textit{sinasuggest}) superficially resemble English words with unfamiliar morphology, evading both Lingua and LLM classification. The remaining cases are English loanwords with Romance or Germanic inflection (\textit{influenciador}, \textit{influenceuse}, \textit{brunchen}, \textit{stressar}), originating from non-English posts where the mixed morphology falls below the language detector's confidence threshold.

\paragraph{The neologism--entity boundary.}
The 134 named entities in the gold standard highlight an inherently
fuzzy boundary. Tokens such as \textit{superstonk} (the subreddit
name that became synonymous with the GameStop movement) and
\textit{barbenheimer} (\textit{Barbie} + \textit{Oppenheimer})
denote specific referents while also exhibiting productive
word-formation processes (compounding, blending), making the
neologism--entity distinction a matter of annotation judgment
rather than a clear-cut category. Game-specific terms from
\textit{Splatoon} (Nintendo, 2015), numbering 15 tokens, \textit{Among Us},
and various crypto projects account for the bulk of entities, and
many are concentrated in one or two subreddits, a pattern
that a cross-subreddit dispersion threshold could help address
(see the Limitations section).

\subsection{Pipeline False Negatives}
\label{sec:discussion:fn}
To estimate recall, we compile a reference list of 103
single-token neologisms documented after 2015 in major
dictionaries and lexicographic sources: Merriam-Webster
additions (2016--2025), Oxford and Collins Words of the Year,
the American Dialect Society Word of the Year, the British
Council ``90~Words'' list, Cambridge Dictionary, the OED,
and Wiktionary, as well as community-maintained documentation
sources such as Know Your Meme
(full list in Appendix~\ref{sec:appendix:recall}).
Of the 103 reference items, 20 are correctly detected;
48 were already attested on Reddit before 2015 and are
correctly excluded; and 2 are excluded from evaluation
as inflected forms of detected base forms.
Loanwords (e.g.\ \textit{mukbang} from Korean) are
excluded, since borrowing falls outside the scope of the
word-formation frameworks adopted in \S\ref{sec:theory}
(see the Limitations section).
Inflected forms of base forms already detected by the pipeline
(e.g.\ \textit{deepfakes} alongside \textit{deepfake}) are
likewise excluded from the false negative count, as the
pipeline's purpose is to identify novel lexical items rather
than to capture every inflectional variant.
The 33 genuine false negatives fall into the following
categories: vocabulary homograph conflicts (17~tokens), where
sparse pre-2015 occurrences in unrelated senses block the
neologism (e.g.\ \textit{rizz} as a character name,
\textit{simp} as a gaming clan); external vocabulary matches
(9~tokens), where WordNet, Wiktionary, or Wikipedia contain the
word under a different meaning (\textit{copium}, \textit{doggo},
\textit{stonks}), a problem closely related to the semantic
shift limitation discussed below; concatenation detection
(3~tokens), where the segmentation module splits the token into
known substrings (e.g.\ \textit{cottagecore}); and the remaining
4~tokens are lost to tokenisation, typo correction, or LLM
misclassification.
Note that \textit{doggo}, used in
\S\ref{sec:theory:mattiello} as a canonical example of
extra-grammatical morphology, is missed precisely because
Wiktionary lists it under its pre-existing adverbial sense.
Recall over the 53 genuinely post-2015
items is 20/53 (37.7\%).
Conversely, \textit{vtuber} (virtual YouTuber), initially flagged
as a typo of \textit{tuber} by SymSpell, was correctly
reintegrated by the frequency threshold mechanism owing to its
41,024 occurrences, illustrating that the reintegration stage
(\S\ref{sec:method:freq}) functions as an effective safety net
for high-frequency neologisms. The main axis for improvement is
therefore conservative refinement of the vocabulary filtering
stage, where type-level matching without sense disambiguation
remains the primary source of false negatives.

\section{Conclusion}
\label{sec:conclusion}

We presented a scalable pipeline for automatic neologism detection
that combines rule-based filtering with multi-model LLM
classification, grounded in grammatical and extra-grammatical
word-formation theory. Applied to 527 million Reddit posts, the
pipeline achieves a 122,031:1 compression ratio, yielding 1,021
candidates of which 599 (58.7\%) are genuine lexical innovations.
Manual analysis of the output reveals a range of productive
word-formation processes, from standard prefixation and
compounding to analogical patterns such as secreted combining
forms, confirming that the pipeline captures theoretically
meaningful variation.
The error analysis indicates that future improvements should target
the earliest filtering stages rather than downstream classification.
The pipeline code, vocabulary compilation scripts, and the annotated
candidate list are available at
\url{https://github.com/DiegoRossini/neologism-pipeline}.

\section*{Ethics Statement}

The pipeline operates on unmoderated social media data and the
resulting candidate list inevitably contains tokens related to
offensive language, sexual content, hate speech, and extremist
ideologies. Their inclusion reflects the lexical productivity of
these domains and does not imply endorsement. All data was processed at the token level; no individual users were identified or tracked.

\section*{Limitations}

The pipeline detects only single-token neologisms. Multi-word
expressions such as \textit{rage bait} or \textit{touch grass} are invisible to the current
architecture because the tokenizer treats each word independently,
and no downstream stage attempts to reassemble multi-token units.
Multi-word expressions are a major vector for lexical innovation in
informal registers, and their absence from the output means the
pipeline systematically underrepresents phrasal coinages.

Neologisms containing numerals or non-alphabetic characters are
likewise excluded: the pattern cleaning stage
(\S\ref{sec:method:pattern}) discards all non-purely-alphabetic
tokens. This filters out an increasingly productive category of
lexical innovation known as algospeak
\citep{steen_you_2023, aleksic_algospeak_2025}, where users
deliberately substitute letters with numbers or symbols to evade
algorithmic content moderation (e.g., \textit{\$3X} for
\textit{sex}, \textit{\$trippers} for \textit{strippers}), as
well as named entities whose orthography includes digits, such as
\textit{4chan}.

The pipeline also cannot detect semantic shifts, where an existing
word acquires a new meaning without any change in form. A
prominent example is \textit{Karen}, a conventional given name
that underwent pejoration on Reddit and Black Twitter during the
mid-2010s to denote an entitled, privileged white woman who
weaponizes her social position. Because \textit{Karen} is already
present in the reference vocabulary as a proper noun, the pipeline
excludes it at the vocabulary filtering stage, and no subsequent
stage is equipped to detect that its usage distribution has changed.

The current instantiation targets English only. While the architecture is modular and transferable, all filtering resources, phonotactic rules, and frequency dictionaries are English-specific, and the LLM prompts are written in English.
Adapting the pipeline to other languages would require substituting
these components and re-evaluating the filtering thresholds.
For morphologically rich languages with extensive inflectional
paradigms, surface-form vocabulary matching may require either
substantially larger observed-form vocabularies or an additional
lemmatization step, though lemmatizing neologisms is itself problematic, since a lemmatizer trained on existing vocabulary may not reliably reduce novel forms to their base.

The foreign language detection stage (\S\ref{sec:method:foreign}) cannot distinguish non-English corpus noise from genuine loanwords entering English. Lexical borrowing is not a word-formation process in either framework adopted in \S\ref{sec:theory} --- neither \citet{stekauer_theory_2002} nor \citet{mattiello_extra-grammatical_2013} include it in their taxonomies, consistent with the position that borrowing and word formation are fundamentally distinct, though interacting, domains \citep{ten_hacken_interaction_2020}. This means that nativised loanwords such as \textit{mukbang} or \textit{hygge}, which are established terms in English, fall outside the pipeline's scope and are excluded either by the language detector or by the reference vocabulary.

As discussed in \S\ref{sec:discussion:fn}, the most consequential
false negatives originate at the tokenization and vocabulary
filtering stages, where tokens attested in pre-2015 data as
probable typos (\textit{stonk}, \textit{monke}) or present in
encyclopaedic sources (\textit{copium}) are silently treated as
known vocabulary and never enter the candidate pool. Two targeted improvements would address recurrent false positive
patterns identified in \S\ref{sec:discussion:errors}. A
cross-subreddit dispersion threshold would complement the raw
frequency threshold by requiring candidates to appear across a
minimum number of distinct subreddits, filtering concatenations such as \textit{datingapp} that accumulate high
frequencies within a single community through repeated
orthographic error rather than deliberate coinage. A
post-classification deduplication step comparing candidates by
edit distance would catch misspellings of neologisms already
captured by the pipeline (e.g., \textit{neurodivegent} alongside
\textit{neurodivergent}).

All filtering thresholds (e.g.\ token length, edit distance,
frequency, language detection confidence) were set based on
preliminary experimentation rather than systematically optimised
on a development set, as the computational cost of a full pipeline
run (50--65 hours) makes exhaustive parameter search impractical.
A systematic sensitivity analysis is left for future work.

\section*{Data and Code Availability}
The pipeline code, vocabulary compilation scripts, and the annotated
candidate list are available at
\url{https://github.com/DiegoRossini/neologism-pipeline}.

\section*{Acknowledgements}
This research was funded by the NCCR Evolving Language,
Swiss National Science Foundation Agreement
No.\ 51NF40\_225146. We thank the anonymous reviewers
for their helpful feedback, Ryan of NoSlang.com for
generously sharing the abbreviation list, and Quirin
W\"{u}rschinger for personal communication regarding
the NeoCrawler.

\section{Bibliographical References}\label{sec:reference}

\bibliographystyle{lrec2026-natbib}
\bibliography{neo_references}

\begin{thebibliography}{8}
\expandafter\ifx\csname natexlab\endcsname\relax\def\natexlab#1{#1}\fi

\bibitem[{Baumgartner et~al.(2020)Baumgartner, Zannettou, Keegan, Squire, and Blackburn}]{baumgartner_pushshift_2020}
Jason Baumgartner, Savvas Zannettou, Brian Keegan, Megan Squire, and Jeremy Blackburn. 2020.
\newblock \href {https://doi.org/10.1609/icwsm.v14i1.7347} {The {P}ushshift {R}eddit dataset}.

\bibitem[{Garbe(2012)}]{symspell}
Wolf Garbe. 2012.
\newblock \href {https://github.com/wolfgarbe/SymSpell} {{SymSpell}: Symmetric delete spelling correction algorithm}.

\bibitem[{{Princeton University}(2011)}]{princeton_wordnet}
{Princeton University}. 2011.
\newblock \href {https://wordnet.princeton.edu/} {{WordNet} 3.1}.

\bibitem[{Stahl(2022)}]{stahl_lingua}
Peter~M. Stahl. 2022.
\newblock \href {https://github.com/pemistahl/lingua-py} {Lingua: The most accurate natural language detection library for python}.

\bibitem[{{Urban Dictionary}(2025)}]{urban_dictionary}
{Urban Dictionary}. 2025.
\newblock \href {https://www.urbandictionary.com/} {{Urban Dictionary} entry database}.
\newblock Scraped and filtered to entries predating 2015.

\bibitem[{{Watchful1}(2025)}]{watchful1_subreddit_2025}
{Watchful1}. 2025.
\newblock \href {https://academictorrents.com/details/1614740ac8c94505e4ecb9d88be8bed7b6afddd4} {Subreddit comments/submissions 2005-06 to 2024-12}.
\newblock Per-subreddit split of the Pushshift Reddit dumps. Available via Academic Torrents.

\bibitem[{{Wikimedia Foundation}(2015{\natexlab{a}})}]{wikimedia_wikipedia_2015}
{Wikimedia Foundation}. 2015{\natexlab{a}}.
\newblock \href {https://dumps.wikimedia.org/} {Wikipedia: {E}nglish article titles dump}.
\newblock Dump dated 2015-01-01.

\bibitem[{{Wikimedia Foundation}(2015{\natexlab{b}})}]{wikimedia_wiktionary_2015}
{Wikimedia Foundation}. 2015{\natexlab{b}}.
\newblock \href {https://dumps.wikimedia.org/} {Wiktionary: {E}nglish edition dump}.
\newblock Dump dated 2015-01-01.

\end{thebibliography}


\begin{thebibliography}{32}
\expandafter\ifx\csname natexlab\endcsname\relax\def\natexlab#1{#1}\fi

\bibitem[{Aleksic(2025)}]{aleksic_algospeak_2025}
Adam Aleksic. 2025.
\newblock \emph{Algospeak: How Social Media Is Transforming the Future of Language}.
\newblock Knopf, New York.

\bibitem[{Arndt-Lappe(2015)}]{arndt-lappe_word-formation_2015}
Sabine Arndt-Lappe. 2015.
\newblock Word-formation and analogy.
\newblock In Peter~O. M{\"u}ller, Ingeborg Ohnheiser, Susan Olsen, and Franz Rainer, editors, \emph{Word-Formation: {An} International Handbook of the Languages of {Europe}}, volume~2, pages 822--841. De Gruyter Mouton, Berlin.

\bibitem[{Booij(2010)}]{booij_construction_2010}
Geert Booij. 2010.
\newblock \emph{Construction Morphology}.
\newblock Oxford University Press, Oxford.

\bibitem[{Brasolin et~al.(2023)Brasolin, Franzini, and Spina}]{brasolin_ti_2023}
Paolo Brasolin, Greta~H. Franzini, and Stefania Spina. 2023.
\newblock "{Ti} blocco perché sei un trollazzo": {Lexical} {Innovation} in {Contemporary} {Italian} in a {Large} {Twitter} {Corpus}.
\newblock \emph{Journal of Italian Linguistics}, 35(2):123--145.

\bibitem[{Cabr\'{e} and de~Yzaguirre(1995)}]{cabre_strategie_1995}
Maria~Teresa Cabr\'{e} and Llu\'{i}s de~Yzaguirre. 1995.
\newblock Strat\'{e}gie pour la d\'{e}tection semi-automatique des n\'{e}ologismes de presse.
\newblock \emph{TTR: Traduction, Terminologie, R\'{e}daction}, 8(2):89--100.

\bibitem[{Cartier(2017)}]{cartier_neoveille_2017}
Emmanuel Cartier. 2017.
\newblock \href {https://aclanthology.org/E17-3024/} {Neoveille, a web platform for neologism tracking}.
\newblock In \emph{Proceedings of the Software Demonstrations of the 15th Conference of the {European} Chapter of the {Association} for {Computational} {Linguistics}}, pages 95--98, Valencia, Spain. Association for Computational Linguistics.

\bibitem[{Dressler(2000)}]{dressler_extragrammatical_2000}
Wolfgang~U. Dressler. 2000.
\newblock Extragrammatical vs. marginal morphology.
\newblock In Ursula Doleschal and Anna~M. Thornton, editors, \emph{Extragrammatical and {Marginal} {Morphology}}, number~12 in {LINCOM} {Studies} in {Theoretical} {Linguistics}, pages 1--10. Lincom Europa, München.

\bibitem[{Falk et~al.(2014)Falk, Bernhard, and G\'{e}rard}]{falk_non_2014}
Ingrid Falk, Delphine Bernhard, and Christophe G\'{e}rard. 2014.
\newblock \href {https://aclanthology.org/L14-1483/} {From non word to new word: Automatically identifying neologisms in {French} newspapers}.
\newblock In \emph{Proceedings of the 9th {International} {Conference} on {Language} {Resources} and {Evaluation} ({LREC} 2014)}, pages 4337--4344, Reykjavik, Iceland. European Language Resources Association (ELRA).

\bibitem[{Grieve et~al.(2018)Grieve, Nini, and Guo}]{grieve_mapping_2018}
Jack Grieve, Andrea Nini, and Diansheng Guo. 2018.
\newblock \href {https://doi.org/10.1177/0075424218793191} {Mapping {Lexical} {Innovation} on {American} {Social} {Media}}.
\newblock \emph{Journal of English Linguistics}, 46(4):293--319.

\bibitem[{Guilbert(1975)}]{guilbert_creativite_1975}
Louis Guilbert. 1975.
\newblock \emph{La cr\'{e}ativit\'{e} lexicale}.
\newblock Langue et Langage. Larousse, Paris.

\bibitem[{Hohenhaus(1998)}]{hohenhaus_non-lexicalizability_1998}
Peter Hohenhaus. 1998.
\newblock Non-lexicalizability as a characteristic feature of nonce word-formation in {English} and {German}.
\newblock \emph{Lexicology}, 4(2):237--280.

\bibitem[{Kerremans et~al.(2018)Kerremans, Prokić, Würschinger, and Schmid}]{kerremans_using_2018}
Daphné Kerremans, Jelena Prokić, Quirin Würschinger, and Hans-Jörg Schmid. 2018.
\newblock \href {https://doi.org/10.1075/pc.00006.ker} {Using data-mining to identify and study patterns in lexical innovation on the web: {The} \textit{{NeoCrawler}}}.
\newblock \emph{Pragmatics \& Cognition}, 25(1):174--200.

\bibitem[{Kerremans et~al.(2012)Kerremans, Stegmayr, and Schmid}]{kerremans_neocrawler_2012}
Daphné Kerremans, Susanne Stegmayr, and Hans-Jörg Schmid. 2012.
\newblock \href {https://doi.org/10.1515/9783110252903.59} {The {NeoCrawler}: {Identifying} and {Retrieving} {Neologisms} from the {Internet} and {Monitoring} {Ongoing} {Change}}.
\newblock In \emph{Current {Methods} in {Historical} {Semantics}}, pages 59--96.

\bibitem[{Klosa-Kückelhaus and Lüngen(2018)}]{klosa_new_2018}
Annette Klosa-Kückelhaus and Harald Lüngen. 2018.
\newblock New {German} words: {Detection}, description, and dictionary entry.
\newblock In \emph{Lexicography in the Digital Age}, pages 559--569. Euralex.

\bibitem[{Körtvélyessy et~al.(2021)Körtvélyessy, Štekauer, and Kačmár}]{kortvelyessy_role_2021}
Lívia Körtvélyessy, Pavol Štekauer, and Pavol Kačmár. 2021.
\newblock \href {https://doi.org/10.1515/ling-2020-0003} {On the role of creativity in the formation of new complex words}.
\newblock \emph{Linguistics}, 59(4):1017--1055.

\bibitem[{Körtvélyessy et~al.(2022)Körtvélyessy, Štekauer, and Kačmár}]{kortvelyessy_creativity_2022}
Lívia Körtvélyessy, Pavol Štekauer, and Pavol Kačmár. 2022.
\newblock \href {https://doi.org/10.1017/9781009053556} {\emph{Creativity in {Word} {Formation} and {Word} {Interpretation}: {Creative} {Potential} and {Creative} {Performance}}}, 1 edition.
\newblock Cambridge University Press.

\bibitem[{Mahler(2020)}]{mahler_lexical_2020}
Taylor Mahler. 2020.
\newblock \href {https://doi.org/10.4000/lexis.4689} {Lexical {Emergence} on {Reddit}}.
\newblock \emph{Lexis -- Journal in English Lexicology}, 16.

\bibitem[{Mattiello(2013)}]{mattiello_extra-grammatical_2013}
Elisa Mattiello. 2013.
\newblock \href {https://doi.org/10.1515/9783110295399} {\emph{Extra-Grammatical Morphology in {English}: {Abbreviations}, {Blends}, {Reduplicatives}, and {Related} {Phenomena}}}.
\newblock Number~82 in Topics in {English} {Linguistics}. De Gruyter Mouton, Berlin.

\bibitem[{Mattiello(2017)}]{mattiello_analogical_2017}
Elisa Mattiello. 2017.
\newblock \href {https://doi.org/10.1515/9783110551419} {\emph{Analogy in Word-formation: {A} Study of {English} Neologisms and Occasionalisms}}.
\newblock Number 309 in Trends in Linguistics. Studies and Monographs. De Gruyter Mouton, Berlin.

\bibitem[{Plag(1999)}]{plag_morphological_1999}
Ingo Plag. 1999.
\newblock \emph{Morphological Productivity: Structural Constraints in {English} Derivation}.
\newblock Mouton de Gruyter, Berlin.

\bibitem[{Renouf(1993)}]{renouf_word_1993}
Antoinette Renouf. 1993.
\newblock A word in time: First findings from dynamic corpus investigation.
\newblock In Jan Aarts, Pieter de~Haan, and Nelleke Oostdijk, editors, \emph{English Language Corpora: Design, Analysis and Exploitation}, pages 279--288. Rodopi, Amsterdam.

\bibitem[{Rey(1976)}]{rey_neologisme_1976}
Alain Rey. 1976.
\newblock N\'{e}ologisme: un pseudo-concept?
\newblock \emph{Cahiers de Lexicologie}, 28(1):3--17.

\bibitem[{Spina et~al.(2024)Spina, Brasolin, and Franzini}]{spina_detecting_2024}
Stefania Spina, Paolo Brasolin, and Greta~H. Franzini. 2024.
\newblock \href {https://doi.org/10.32714/ricl.13.01.07} {Detecting emerging vocabulary in a large corpus of {Italian} tweets}.
\newblock \emph{Research in Corpus Linguistics}, 13(1):139--170.

\bibitem[{Steen et~al.(2023)Steen, Yurechko, and Klug}]{steen_you_2023}
Ella Steen, Kathryn Yurechko, and Daniel Klug. 2023.
\newblock \href {https://doi.org/10.1177/20563051231194586} {You can (not) say what you want: Using algospeak to contest and evade algorithmic content moderation on {TikTok}}.
\newblock \emph{Social Media + Society}, 9(3).

\bibitem[{{\v{S}}tekauer(2001)}]{stekauer_fundamental_2001}
Pavol {\v{S}}tekauer. 2001.
\newblock Fundamental principles of an onomasiological theory of {English} word-formation.
\newblock \emph{Onomasiology Online}, 2:1--42.

\bibitem[{ten Hacken and Panocov\'{a}(2020)}]{ten_hacken_interaction_2020}
Pius ten Hacken and Ren\'{a}ta Panocov\'{a}, editors. 2020.
\newblock \href {https://doi.org/10.1515/9781474448215} {\emph{The Interaction of Borrowing and Word Formation}}.
\newblock Edinburgh University Press, Edinburgh.

\bibitem[{Tomaszewska et~al.(2025)Tomaszewska, Czerski, Żuk, and Ogrodniczuk}]{tomaszewska_neon_2025}
Aleksandra Tomaszewska, Dariusz Czerski, Bartosz Żuk, and Maciej Ogrodniczuk. 2025.
\newblock \href {https://doi.org/10.48550/arXiv.2505.15426} {{NeoN}: {A} {Tool} for {Automated} {Detection}, {Linguistic} and {LLM}-{Driven} {Analysis} of {Neologisms} in {Polish}}.
\newblock ArXiv:2505.15426 [cs].

\bibitem[{Würschinger(2021)}]{wurschinger_social_2021}
Quirin Würschinger. 2021.
\newblock \href {https://doi.org/10.3389/frai.2021.648583} {Social {Networks} of {Lexical} {Innovation}. {Investigating} the {Social} {Dynamics} of {Diffusion} of {Neologisms} on {Twitter}}.
\newblock \emph{Frontiers in Artificial Intelligence}, 4:648583.

\bibitem[{Zwicky and Pullum(1987)}]{zwicky_plain_1987}
Arnold~M. Zwicky and Geoffrey~K. Pullum. 1987.
\newblock \href {https://doi.org/10.3765/bls.v13i0.1817} {Plain morphology and expressive morphology}.
\newblock In \emph{Proceedings of the Thirteenth Annual Meeting of the {Berkeley Linguistics Society}}, pages 330--340.

\bibitem[{Štekauer(1998)}]{stekauer_onomasiological_1998}
Pavol Štekauer. 1998.
\newblock \href {https://doi.org/10.1075/sfsl.46} {\emph{An {Onomasiological} {Theory} of {English} {Word}-{Formation}}}, volume~46 of \emph{Studies in {Functional} and {Structural} {Linguistics}}.
\newblock John Benjamins Publishing Company, Amsterdam.

\bibitem[{Štekauer(2002)}]{stekauer_theory_2002}
Pavol Štekauer. 2002.
\newblock \href {https://doi.org/10.1080/07268600120122571} {On the {Theory} of {Neologisms} and {Nonce}-formations}.
\newblock \emph{Australian Journal of Linguistics}, 22(1):97--112.

\bibitem[{Štekauer(2005)}]{den_dikken_onomasiological_2005}
Pavol Štekauer. 2005.
\newblock \href {https://doi.org/10.1007/1-4020-3596-9_9} {Onomasiological {Approach} to {Word}-{Formation}}.
\newblock In Marcel Den~Dikken, Liliane Haegeman, Joan Maling, Guglielmo Cinque, Carol Georgopoulos, Jane Grimshaw, Michael Kenstowicz, Hilda Koopman, Howard Lasnik, Alec Marantz, John~J. McCarthy, Ian Roberts, Pavol Štekauer, and Rochelle Lieber, editors, \emph{Handbook of {Word}-{Formation}}, volume~64, pages 207--232. Springer Netherlands, Dordrecht.
\newblock Series Title: Studies in Natural Language and Linguistic Theory.

\end{thebibliography}

\section{Language Resource References}
\label{lr:ref}
\bibliographystylelanguageresource{lrec2026-natbib}
\bibliographylanguageresource{languageresource}

\appendix

\section{Prompt Templates}
\label{sec:appendix:prompts}

Both prompts are used identically across all four models (Qwen 72B, LLaMA 70B, Mistral Large 123B, and Claude Haiku).

\paragraph{Multi-token prompt (primary pass, 10 tokens per call).}

\begin{footnotesize}
\begin{verbatim}
TASK: Classify each token
into ONE category.

ENTITY - Pure proper nouns only
  (real/fictional): people, characters,
  companies, brands, products, games,
  movies, places, apps
  Examples: elon, pikachu, google,
  iphone, fortnite, reddit, tokyo

NEOLOGISM - New English words, slang,
  OR words derived from proper nouns
  Examples: doomscrolling, ghosting,
  rizz, bussin, adulting, covidiot,
  youtuber, redditor, trumpian,
  instagrammable, uberize, googlable

FOREIGN - Non-English words
  Examples: além, anspielung,
  yapmyorum, además

NONE - Usernames, typos, programming
  terms, unclear words

CRITICAL RULES:
1. Derived forms are NEOLOGISM
   (youtuber -> NEOLOGISM,
    youtube -> ENTITY)
2. When uncertain, classify as NONE
3. Use the context and subreddit
   to understand usage

TOKENS:
TOKEN: <token_1>
  context_1 (r/<subreddit>): "<text>"
  context_2 (r/<subreddit>): "<text>"
  context_3 (r/<subreddit>): "<text>"
TOKEN: <token_2>
  context_1 (r/<subreddit>): "<text>"
  ...

OUTPUT:
One classification per line as
TOKEN:LABEL (ENTITY, NEOLOGISM,
FOREIGN, or NONE).
No explanations.
\end{verbatim}
\end{footnotesize}

\paragraph{Single-token prompt (retry pass for failed tokens).}

\begin{footnotesize}
\begin{verbatim}
Classify this token into ONE
category: ENTITY, NEOLOGISM,
FOREIGN, or NONE.

ENTITY - Pure proper nouns only
  (real/fictional): people,
  characters, companies, brands,
  products, games, movies, places,
  apps
NEOLOGISM - New English words, slang,
  OR words derived from proper nouns
  (youtuber, trumpian,
   instagrammable)
FOREIGN - Non-English words
NONE - Usernames, typos, programming
  terms, unclear words

TOKEN: <token>
  context_1 (r/<subreddit>): "<text>"
  context_2 (r/<subreddit>): "<text>"
  context_3 (r/<subreddit>): "<text>"

Answer with ONLY the label:
<token>:LABEL
\end{verbatim}
\end{footnotesize}

\section{Recall Reference List}
\label{sec:appendix:recall}

Table~\ref{tab:recall_1} and Table~\ref{tab:recall_2} list the
103 single-token neologisms used for the recall evaluation in
\S\ref{sec:discussion:fn}.

\paragraph{Status labels.}
TP = detected by the pipeline (true positive);
FN = genuine false negative (post-2015, missed by pipeline);
pre-15 = correctly excluded (attested on Reddit before 2015);
excl.\ = excluded from evaluation (inflected form of a
detected base form).

\paragraph{Source abbreviations.}
MW = Merriam-Webster;
KYM = Know Your Meme;
BC90 = British Council 90 Words;
ADS = American Dialect Society;
Collins = Collins Dictionary;
Cambridge = Cambridge Dictionary;
Oxford WOTY = Oxford Word of the Year;
UrbanDict = Urban Dictionary;
Dictionary.com = Dictionary.com;
Aesth.~Wiki = Aesthetics Wiki.\footnote{Source base URLs:
Merriam-Webster: \url{https://www.merriam-webster.com};
Know Your Meme: \url{https://knowyourmeme.com};
British Council 90 Words: \url{https://www.britishcouncil.org};
American Dialect Society: \url{https://www.americandialect.org};
Collins Dictionary: \url{https://www.collinsdictionary.com};
Cambridge Dictionary: \url{https://dictionary.cambridge.org};
Oxford Word of the Year: \url{https://languages.oup.com/word-of-the-year};
Urban Dictionary: \url{https://www.urbandictionary.com};
Dictionary.com: \url{https://www.dictionary.com};
Wiktionary: \url{https://en.wiktionary.org};
Aesthetics Wiki: \url{https://aesthetics.fandom.com}.
The full reference list with per-word
verification URLs is available in the
project repository.}

\begin{table*}[t]
\centering
\footnotesize
\begin{tabular*}{\textwidth}{@{\extracolsep{\fill}}llll@{}}
\toprule
\textbf{Word} & \textbf{Year} & \textbf{Source} & \textbf{Status} \\
\midrule
\textit{doomscroll} & 2020 & MW 2023 & TP \\
\textit{doomscrolling} & 2020 & MW 2023 & TP \\
\textit{deepfake} & 2017 & MW 2023; BC90 & TP \\
\textit{deepfakes} & 2017 & MW 2023 & excl. \\
\textit{finsta} & 2017 & MW 2023 & FN \\
\textit{edgelord} & 2016 & MW 2023; BC90 & pre-15 \\
\textit{copypasta} & 2016 & MW 2023 & pre-15 \\
\textit{clickbait} & 2015 & MW 2018 & pre-15 \\
\textit{subtweet} & 2015 & MW 2018 & pre-15 \\
\textit{doxing} & 2015 & MW 2017 & pre-15 \\
\textit{doxxing} & 2015 & MW 2023 & pre-15 \\
\textit{ghosting} & 2017 & MW 2017 & pre-15 \\
\textit{catfishing} & 2015 & MW 2023 & pre-15 \\
\textit{copium} & 2020 & Collins; MW & FN \\
\textit{hopium} & 2020 & Collins & pre-15 \\
\textit{shitposting} & 2017 & Wiktionary & pre-15 \\
\textit{shitpost} & 2017 & Wiktionary & pre-15 \\
\textit{rizz} & 2023 & MW 2023; Oxford WOTY 2023; BC90 & FN \\
\textit{simp} & 2019 & MW 2023 & FN \\
\textit{simping} & 2019 & MW 2025 & FN \\
\textit{stan} & 2017 & MW 2019 & pre-15 \\
\textit{stanning} & 2017 & MW 2019 & pre-15 \\
\textit{sealioning} & 2017 & Collins & pre-15 \\
\textit{doggo} & 2017 & MW 2023 & FN \\
\textit{birb} & 2017 & KYM & FN \\
\textit{chonk} & 2018 & KYM & FN \\
\textit{chonky} & 2018 & KYM & FN \\
\textit{poggers} & 2017 & KYM & FN \\
\textit{stonks} & 2021 & KYM & FN \\
\textit{thicc} & 2017 & KYM & FN \\
\textit{updoot} & 2016 & KYM & TP \\
\textit{yeet} & 2018 & MW 2023 & pre-15 \\
\textit{yeeted} & 2018 & MW 2023 & FN \\
\textit{sussy} & 2021 & KYM & FN \\
\textit{bussin} & 2021 & MW 2023 & pre-15 \\
\textit{skibidi} & 2023 & Cambridge 2025 & FN \\
\textit{delulu} & 2023 & Cambridge 2025 & FN \\
\textit{uwu} & 2017 & KYM & pre-15 \\
\textit{smol} & 2016 & KYM & FN \\
\textit{blorbo} & 2022 & KYM & TP \\
\textit{enshittification} & 2023 & ADS WOTY 2023 & TP \\
\textit{enshittify} & 2023 & Wiktionary & FN \\
\textit{touchgrass} & 2021 & MW 2024 & FN \\
\textit{blockchain} & 2016 & MW 2018 & pre-15 \\
\textit{cryptocurrency} & 2017 & MW 2018 & pre-15 \\
\textit{bitcoin} & 2016 & MW 2016 & pre-15 \\
\textit{chatbot} & 2017 & MW 2018 & pre-15 \\
\textit{ransomware} & 2017 & MW 2018 & pre-15 \\
\textit{deepfaked} & 2019 & Wiktionary & FN \\
\textit{vtuber} & 2020 & Wiktionary & TP \\
\textit{hodl} & 2017 & Wiktionary & pre-15 \\
\textit{defi} & 2020 & Wiktionary & pre-15 \\
\bottomrule
\end{tabular*}
\caption{Recall reference list (1/2). "Year" indicates when the source documented the word, not the year of coinage.}\label{tab:recall_1}
\end{table*}

\begin{table*}[t]
\centering
\footnotesize
\begin{tabular*}{\textwidth}{@{\extracolsep{\fill}}llll@{}}
\toprule
\textbf{Word} & \textbf{Year} & \textbf{Source} & \textbf{Status} \\
\midrule
\textit{altcoin} & 2017 & Wiktionary & pre-15 \\
\textit{memecoin} & 2021 & Wiktionary & pre-15 \\
\textit{stablecoin} & 2020 & Wiktionary & pre-15 \\
\textit{rugpull} & 2021 & Wiktionary & FN \\
\textit{rugpulled} & 2021 & Wiktionary & FN \\
\textit{wokeism} & 2019 & Wiktionary & TP \\
\textit{wokeness} & 2019 & Wiktionary & FN \\
\textit{trumpism} & 2016 & Wiktionary & TP \\
\textit{deplatform} & 2018 & Wiktionary & TP \\
\textit{deplatformed} & 2018 & Wiktionary & excl. \\
\textit{deplatforming} & 2018 & Wiktionary & TP \\
\textit{mansplaining} & 2015 & MW 2018 & pre-15 \\
\textit{manspreading} & 2015 & MW 2016 & pre-15 \\
\textit{whataboutism} & 2017 & MW 2019 & pre-15 \\
\textit{incel} & 2018 & Collins WOTY 2018; BC90 & pre-15 \\
\textit{incels} & 2018 & Collins WOTY 2018 & pre-15 \\
\textit{blackpill} & 2018 & Wiktionary & FN \\
\textit{blackpilled} & 2018 & Wiktionary & TP \\
\textit{redpilled} & 2016 & Wiktionary & pre-15 \\
\textit{breadcrumbing} & 2018 & Wiktionary & TP \\
\textit{situationship} & 2022 & Oxford WOTY 2023; BC90 & pre-15 \\
\textit{allyship} & 2018 & MW 2019 & pre-15 \\
\textit{covidiot} & 2020 & Collins & TP \\
\textit{quarantini} & 2020 & Wiktionary & FN \\
\textit{longcovid} & 2020 & Wiktionary & TP \\
\textit{superspreader} & 2020 & MW 2020 & FN \\
\textit{doomscroller} & 2020 & MW 2023 & TP \\
\textit{covfefe} & 2017 & Wiktionary & TP \\
\textit{infodemic} & 2020 & Wiktionary & FN \\
\textit{deadname} & 2018 & MW 2023 & FN \\
\textit{deadnaming} & 2018 & MW 2023 & FN \\
\textit{genderfluid} & 2016 & MW 2018 & pre-15 \\
\textit{demisexual} & 2018 & Wiktionary & pre-15 \\
\textit{neurodivergent} & 2020 & MW 2023 & pre-15 \\
\textit{adulting} & 2016 & MW 2017; BC90 & pre-15 \\
\textit{cottagecore} & 2020 & Dictionary.com & FN \\
\textit{goblincore} & 2020 & Dictionary.com & TP \\
\textit{darkcore} & 2020 & Aesth.~Wiki & pre-15 \\
\textit{tradwife} & 2020 & Cambridge 2025 & TP \\
\textit{sponcon} & 2018 & Wiktionary & FN \\
\textit{finfluencer} & 2020 & Wiktionary & TP \\
\textit{hygge} & 2016 & Oxford WOTY 2016 & pre-15 \\
\textit{glamping} & 2016 & MW 2016 & pre-15 \\
\textit{athleisure} & 2016 & MW 2016 & pre-15 \\
\textit{shadowban} & 2022 & MW 2024 & pre-15 \\
\textit{jawnz} & 2023 & UrbanDict & pre-15 \\
\textit{longhauler} & 2020 & Wiktionary & FN \\
\textit{rawdogging} & 2024 & Wiktionary & pre-15 \\
\textit{brainrot} & 2024 & Oxford WOTY 2024 & FN \\
\textit{airdrop} & 2017 & Wiktionary & pre-15 \\
\textit{gaslighting} & 2016 & MW WOTY 2022 & pre-15 \\
\bottomrule
\end{tabular*}
\caption{Recall reference list (2/2). "Year" indicates when the source documented the word, not the year of coinage.}\label{tab:recall_2}
\end{table*}

\end{document}